\pdfoutput=1

\documentclass[JACIII,paper]{jaciiiarticle}
\usepackage{graphicx}
\usepackage{epstopdf}
\usepackage{epsfig}
\usepackage{array}
\usepackage{multirow}
\usepackage{float}
\usepackage[caption=false]{subfig}
\usepackage{color}

\begin{document}

\pagestyle{jaciiistyle}

\title{Image Crowd Counting Using Convolutional Neural Network and Markov Random Field}
\author{Kang Han, Wanggen Wan, Haiyan Yao, and Li Hou}
\address{School of Communication and Information Engineering, Shanghai University, Shanghai, China\\
    Institute of Smart City, Shanghai University, Shanghai, China\\
         E-mail: shuye.ys@gmail.com}
\markboth{Han Kang. et al.}{Image Crowd Counting Using CNN and MRF}
\maketitle

\begin{abstract}
\noindent Abstract. 
	In this paper, we propose a method called Convolutional Neural Network-Markov Random Field (CNN-MRF) to estimate the crowd count in a still image. We first divide the dense crowd visible image into overlapping patches and then use a deep convolutional neural network to extract features from each patch image, followed by a fully connected neural network to regress the local patch crowd count. Since the local patches have overlapping portions, the crowd count of the adjacent patches has a high correlation. We use this correlation and the Markov random field to smooth the counting results of the local patches. Experiments show that our approach significantly outperforms the state-of-the-art methods on UCF and Shanghaitech crowd counting datasets. Code available on GitHub https://github.com/hankong/crowd-counting.
\end{abstract}


\section{Introduction}
In modern society, more and more people gather to live in the city. This lifestyle provides convenience for people's lives and improves the utilization rate of urban public resources. Meanwhile, a large number of people living in cities also led to urban congestion problems. This crowding phenomenon can be observed at traffic junctions, airports, stations, stadiums and other public places. Many accidents caused by overcrowding have led to many deaths, such as massive stampede happened in Shanghai Bund in 2015, where 36 persons died and 49 persons were injured. Therefore, automatic crowd density estimation and early warning for overcrowding are important to prevent such tragedies from happening again.

\indent Existing methods for people density estimation and counting of the crowd can be divided into two main categories: direct and indirect approaches \cite{saleh2015recent}. The main idea of the direct approach (also called object detection based) is to detect and segment each individual in crowd scenes to get the total count, while the indirect approach (also called feature based) takes an image as a whole and extracts some features, and then get the final count through the regression analysis of the extracted features. The advantage of the direct method is that people or head detection have been widely studied and applied, these methods can be easily adapted to the crowd of few tens \cite{dalal2005histograms,stewart2016end,yasuoka2016simulation}. However, a crowd of more than hundreds does not have a well-defined shape as a single object does, so the direct detection method is not applicable. In this scene, the indirect methods are generally more reliable and efficient, since the overall features of a crowd image are easier to obtain and have a stronger correlation with the number of people. Some surveys of these methods can be seen in \cite{junjie2017a}.\par 

In this paper, we focus on the indirect method to estimate the crowd count of more than hundreds.  Like other computer vision tasks, features extraction is the first step and the most important step in the crowd counting problem. Many hand-crafted computer vision features have been used to represent the density of the crowd, such as Local Binary Patterns (LBP) \cite{ojala2002multiresolution}, Scale Invariant Feature Transform (SIFT) \cite{lowe2004distinctive} and Histogram of Oriented Gradient (HOG) \cite{dalal2005histograms}. However, due to the variation of viewpoint, scene and crowd count, these hand-crafted features can not represent the crowd density discriminatively. In recent years, with the advancement of deep learning in computer vision, some researchers try to apply deep learning to crowd density estimation and achieved state-of-the-art results \cite{zhang2016single}. Different with the hand-crafted features extraction which follows the certain steps, deep learning can automatically learn the features from the data. Following the certain steps means that the features will not be better with the increase of the data, while deep learning can learn more discriminative features from abundant data. \par

\indent Some auxiliary methods are also used to improve the accuracy of the crowd density estimation. For example, in the crowd of more than hundreds, the density of crowd is continuously gradient, so MRF can be used to smooth the counting between adjacent patches \cite{idrees2013multi}.\par

\indent Inspired by the superior feature representation of deep learning and the smoothness of MRF, we propose a CNN and MRF based framework for the problem of people counting in still images. Firstly, we divide the image into patches with overlaps and use a pre-trained CNN model to extract deep features from each overlapping patch, followed by a fully connected deep neural network to regress the patch people count. Finally, we use MRF to smooth the counts of adjacent patches in order to make the counts closer to the true value. The reason is that the overlap between the adjacent patches leads to the people count of adjacent patches keep a certain consistency.\par

\indent The rest of the paper is organized as follows. In Section 2, we briefly review the related work of crowd density estimation and counting. And then the system architecture and the implement detail of the proposed approach will be illustrated in Section 3, followed by the experiment in Section 4. Finally, Section 5 concludes our paper.

\section{Related work}
\textbf{Direct method}. Li et al \cite{li2008estimating} proposed a people count estimation method combining the foreground segmentation and the head-shoulder detection. A Mosaic Image Difference based foreground segmentation algorithm is performed first to detect active areas, and then a head-shoulder detection algorithm is utilized to detect heads and count the number from the detected foreground areas. Cheriyadat et al \cite{cheriyadat2008detecting} presented an object detection system based on coherent motion region detection for counting and locating objects in the presence of high object density and inter-object occlusions. They used the locations of tracked low-level feature points to construct all possible coherent-motion-regions and chose a good disjoint set of coherent motion regions representing individual objects using a greedy algorithm. Brostow et al \cite{brostow2006unsupervised} described an unsupervised data-driven Bayesian clustering algorithm to detect individual entities. The method tracked simple image features and probabilistically group them into clusters representing independently moving entities.\par

\indent \textbf{Indirect method}. Lempitsky et al \cite{lempitsky2010learning} used dense SIFT features and  Maximum Excess over SubArrays distance as a loss function to train a regression model on randomly selected patches. In order to adapt to the change of the crowd density and perspective, Zhang et al \cite{zhang2016single} proposed a Multi-column CNN architecture to map the image to its crowd density map. The network structure included three parallel CNN with different sizes to extract features from different scales. The features learned by each column CNN were adaptive to variations in people/head size due to perspective effect or image resolution. Shang et al \cite{shang2016end} proposed an end-to-end CNN architecture that directly maps the whole image to the counting result. A pre-trained GoogLeNet model was used to extract high-level deep features and the long-short time memory (LSTM) decoders for the local count and fully connected layers for the final count. A cross-scene crowd counting architecture was proposed by Zhang et al \cite{zhang2015cross}. Firstly, they trained a deep CNN with two related learning objectives, crowd density, and crowd count. And then a data-driven method was introduced to fine-tune the learned CNN to an unseen target scene, where the training samples similar to the target scene were retrieved from the training scenes for fine-tuning.\par

Some researchers tried to combine two methods to get a more accurate estimate count. Idrees et al \cite{idrees2013multi} proposed a multi-source multi-scale counting method to compute an estimation of the number of individuals presented in an extremely dense crowd visible in a single image. This method combined the detection approach (low confidence head detection) and the features regression approach (repetition of texture element and frequency-domain analysis) to solve the perspective, occlusion, clutter and few pixels per person problems in a crowd of more than hundreds. Then the spatial consistency constraints are employed by MRF to smooth the counting between adjacent patches. Rodriguez et al. \cite{rodriguez2011density} addressed the problem of person detection and tracking in crowded video scenes. They explored constraints imposed by the crowd density and formulate person detection as the optimization of a joint energy function combining crowd density estimation and the localization of individual people.
\begin{figure}[ht]
    \centering
    \subfloat[]{\includegraphics[width=3.5cm,height=3.0cm]{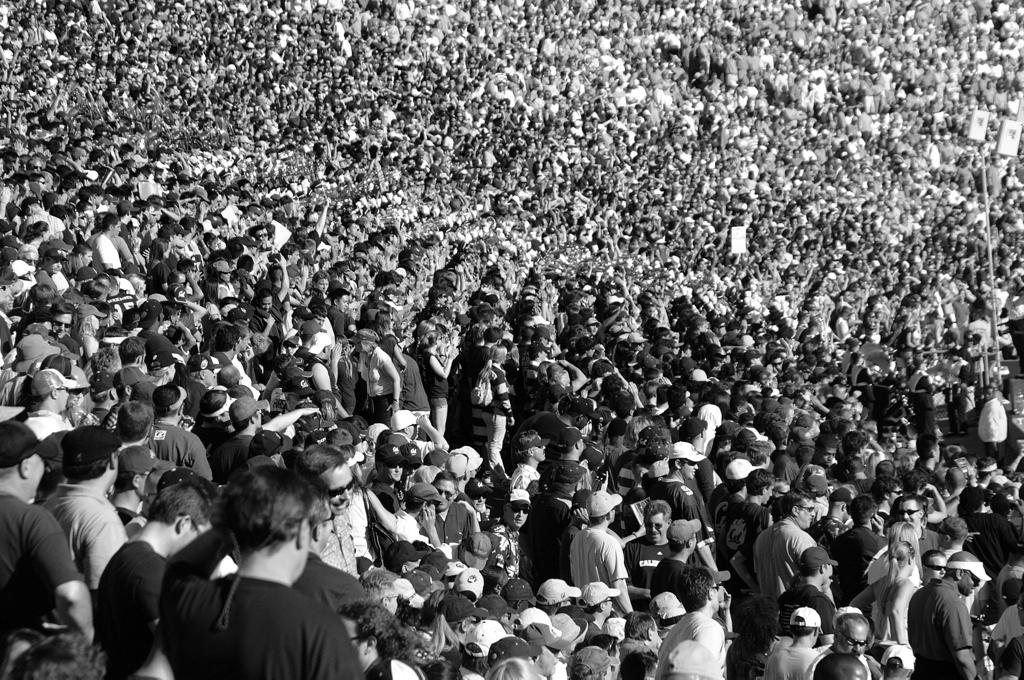}} \quad
    \subfloat[]{\includegraphics[width=3.5cm,height=3.0cm]{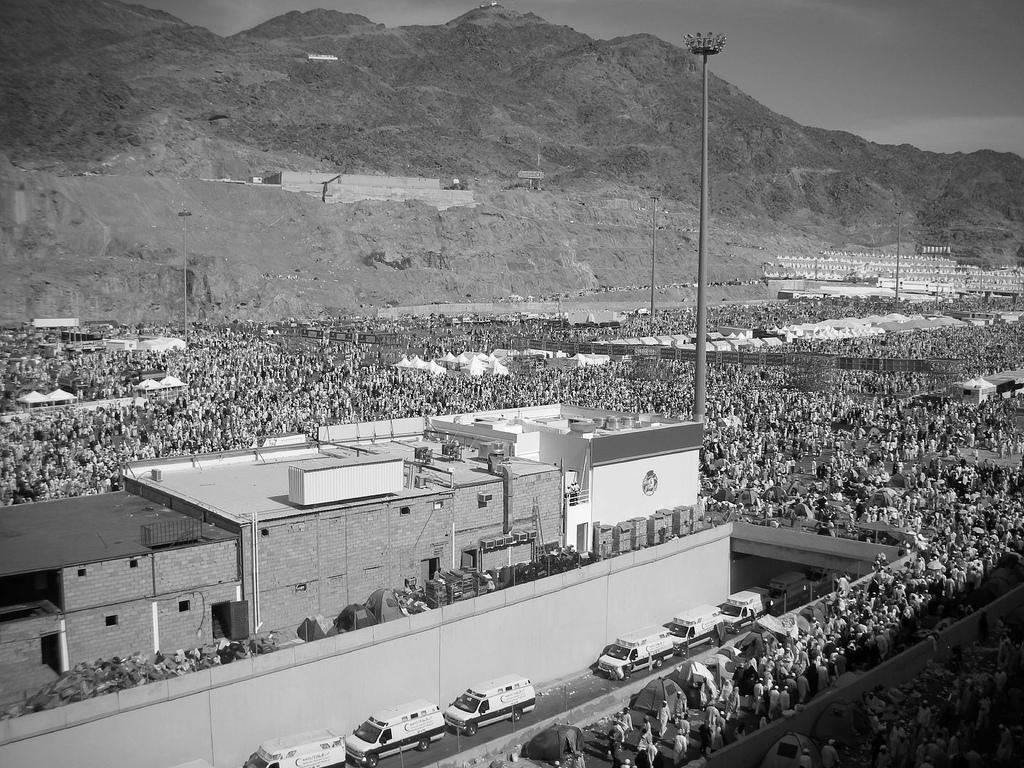}} \quad
    \caption{Some example images from UCF crowd dataset. The crowd density become sparse with the view from far to near in image (a). The crowd density of the buildings and mountains is 0 in the image (b).}
    \label{fig:exp}
\end{figure}

\begin{figure*}[t]
  \centering
  \includegraphics[width=0.95\linewidth]{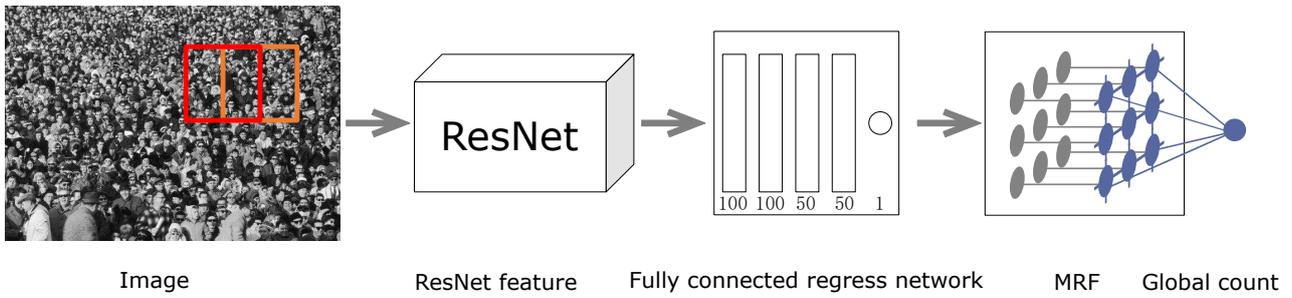}
  \caption{The overview of our proposed CNN-MRF crowd counting method. The proposed method contains three parts, a pre-trained deep residual network to extract features, a fully connected neural network for count regress and a MRF to smooth the counting results of the local patches. The final count is obtained by the sum of the patches count.}
  \label{fig:system}
\end{figure*}

\section{Proposed Method}
For dense crowd images, the distribution of crowd density is rarely uniform due to changes in perspective and scene. Some example images can be seen in Fig. \ref{fig:exp}. Therefore, it is unreasonable to count the crowd by taking the image as an entirety. So our framework adapted the divide-count-sum strategy. The images are firstly divided into patches, then a regression model is learned to map the image patch to the local count. Finally, the global image count is computed as the total sum over these patches. The image segmentation has two advantages: Firstly, in the small image patches, the crowd density is approximately uniform distribution. Secondly, the segmentation of image increases the number of training data for the regression model. The above advantages allow us to train a more robust regression model. \par
\indent Even though the distribution of crowd density is not uniform, the overall crowd density distribution is continuous. This means that the density of adjacent image patches should be similar. Furthermore, we divide the image with overlaps, which enhances the association between image patches. The Markov random field is used to smooth the estimation count between overlapping image patches to compensate for the possible estimation errors of the image patches and to bring the overall result closer to the true density distribution. The overview of the proposed method can be seen as Fig. \ref{fig:system}.

\subsection{Patches counting}
We use a pre-trained deep residual network to extract features from image patches and a fully connected neural network to learn a map from the above features to the local count. The features learned from deep convolutional networks have been used for many computer vision tasks such as image recognition, object detection, and image segmentation \cite{he2015deep}. This indicates that the features learned from the deep convolutional network are universal to many computer vision tasks. With the increase in the number of network layers, the representation ability of the learned features becomes stronger. However, a deeper model means that more data is needed for training. The existing crowd counting datasets are not large enough to train a very deep convolutional neural network from scratch. So we use a pre-trained deep residual network to extract features from image patches. The deep residual network was proposed by He et al. \cite{he2015deep}. Their method addressed the degradation problem by reformulating the layers as learning residual functions with reference to the layer inputs, instead of learning unreferenced functions. We employ the residual network, which is trained on ImageNet dataset for image classification task, to extract the deep features to represent the density of the crowd. This pre-trained CNN network created a residual item for every three convolution layer to bring the layer of the network to 152. We resize the image patches to the size of 224 $\times$ 224 as the input of the model and extract the output of the \emph{fc1000} layer to get the 1000 dimensional features.\par

The features are then used to train 5 layers fully connected neural network. The network's input is 1000-dimensional, and
the number of neurons in the network is given by 100-100-50-50-1. The network's output is the local crowd count. The learning task of the fully connected neural network is to minimize the mean squared error of the training patches, which can be written as:
\begin{equation}
    Loss = \sqrt{\frac{1}{M}\sum_{i=1}^{M}(c_g - c_r)^2}
\end{equation}
where $M$ is the number of the training image patches and \(c_g\) and \(c_r\) are the ground truth count and the regression count of the image patches, respectively.

\subsection{Images counting}
Due to the overlapping of the adjacent image patches, there is a high correlation between adjacent local people count. This correlation can be used by the Markov random field to smooth the estimation count between adjacent image patches. As previously analyzed, the people count of adjacent images patches is generally similar and may change dramatically at some places due to buildings or other objects in the scene. This characteristic can be well modeled by the Markov random field. Formally, the Markov random field framework for the crowd counting can be defined as follows (we follow the notation in \cite{felzenszwalb2006efficient}). Let \(P\) be the set of patches in an image and \(C\) be a possible set of counts. A counting \(c\) assigns a count \(c_p \in C\) to each patch \(p \in P\). The quality of a counting is given by an energy function:
\begin{equation}
    E(c) = \sum_{p \in P}D_p(c_p) + \sum_{(p, q) \in N}V(c_p - c_q)
\end{equation}
where \(N\) are the (undirected) edges in the four-connected image patch graph. \(D_p(c_p)\) is the cost of assigning count \(c_p\) to patch \(p\), and is referred to as the data cost. \(V(c_p - c_q)\) measures the cost of assigning count \(c_p\) and \(c_q\) to two neighboring patch, and is normally referred to as the discontinuity cost. \par
\begin{figure*}[!ht]
    \centering
    \captionsetup[subfigure]{labelformat=empty}
    \subfloat[Images]{
        \begin{tabular}{@{}c@{}}
            \includegraphics[width=3.5cm,height=2.8cm]{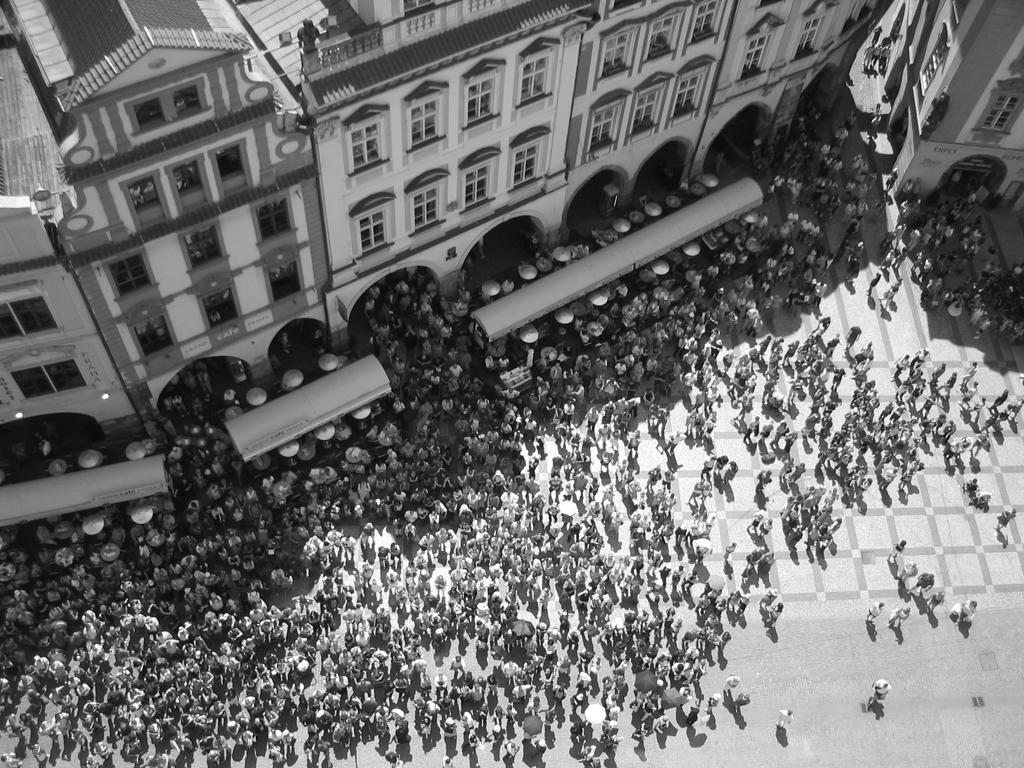} \\
            \includegraphics[width=3.5cm,height=2.8cm]{./fig/density_SHT}
        \end{tabular}
    }
    \subfloat[Ground Truth]{
        \begin{tabular}{@{}c@{}}
            \includegraphics[width=3.5cm,height=2.8cm]{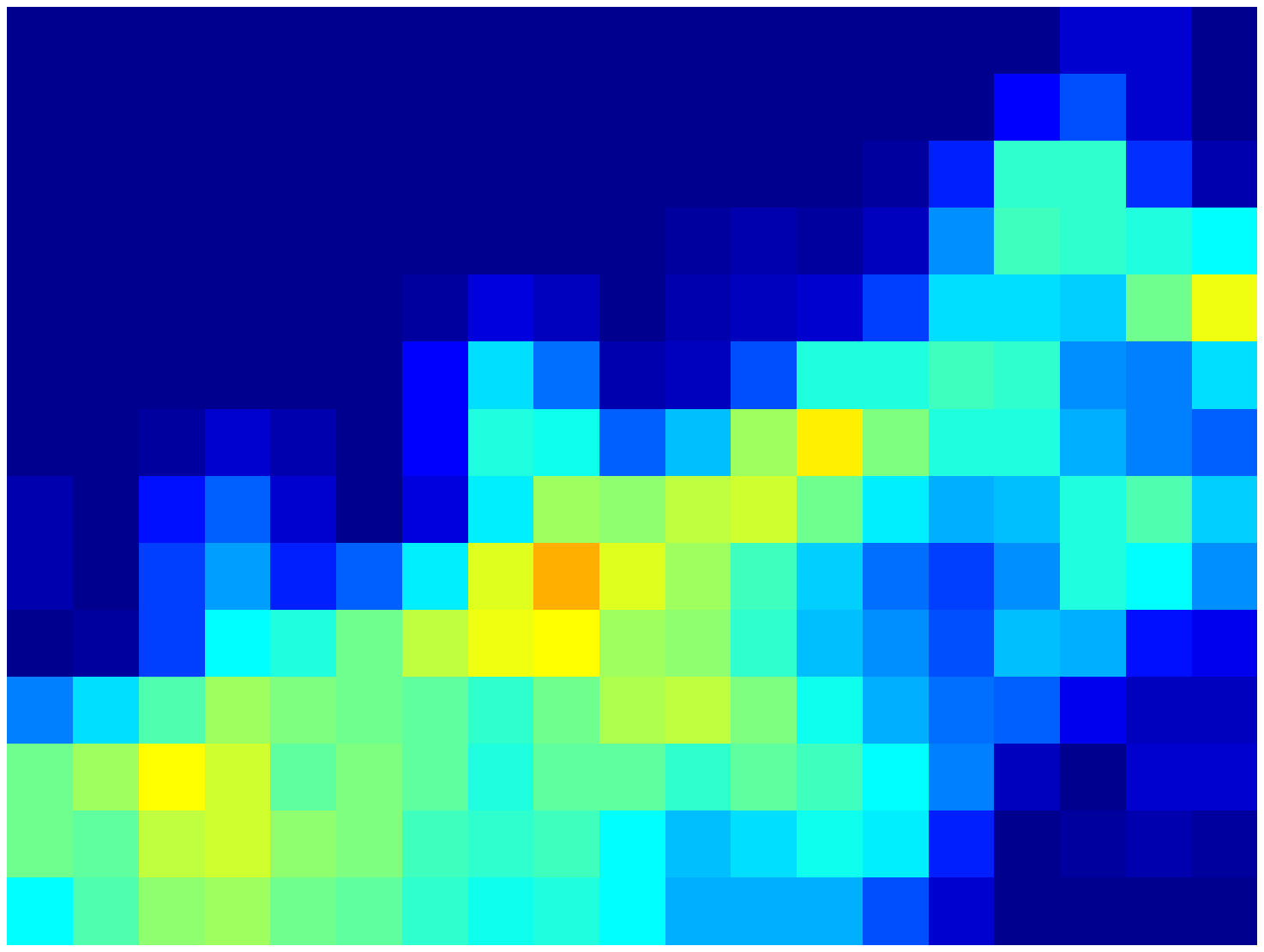} \\
            \includegraphics[width=3.5cm,height=2.8cm]{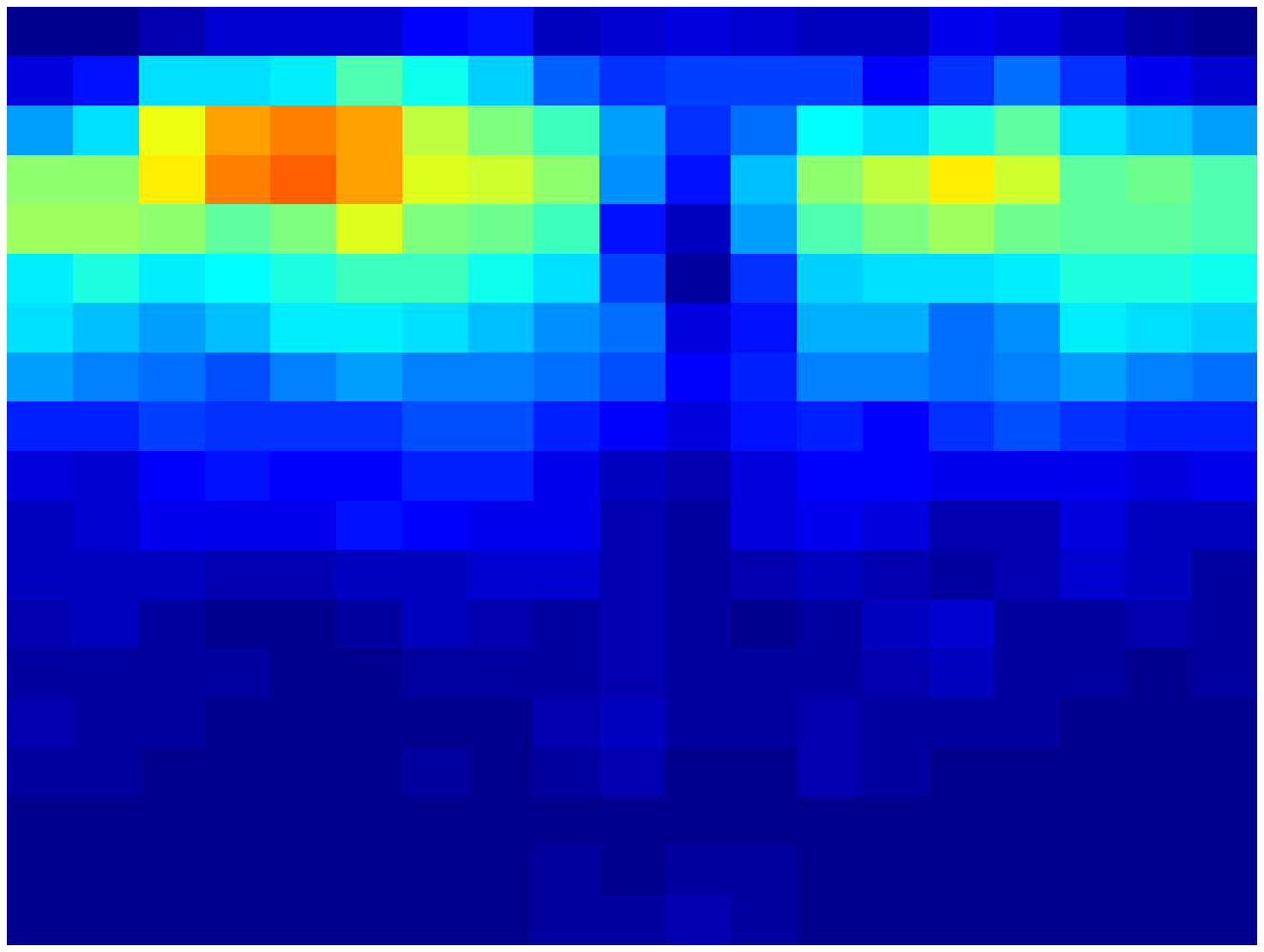}
        \end{tabular}
    }
    \subfloat[Before MRF]{
        \begin{tabular}{@{}c@{}}
            \includegraphics[width=3.5cm,height=2.8cm]{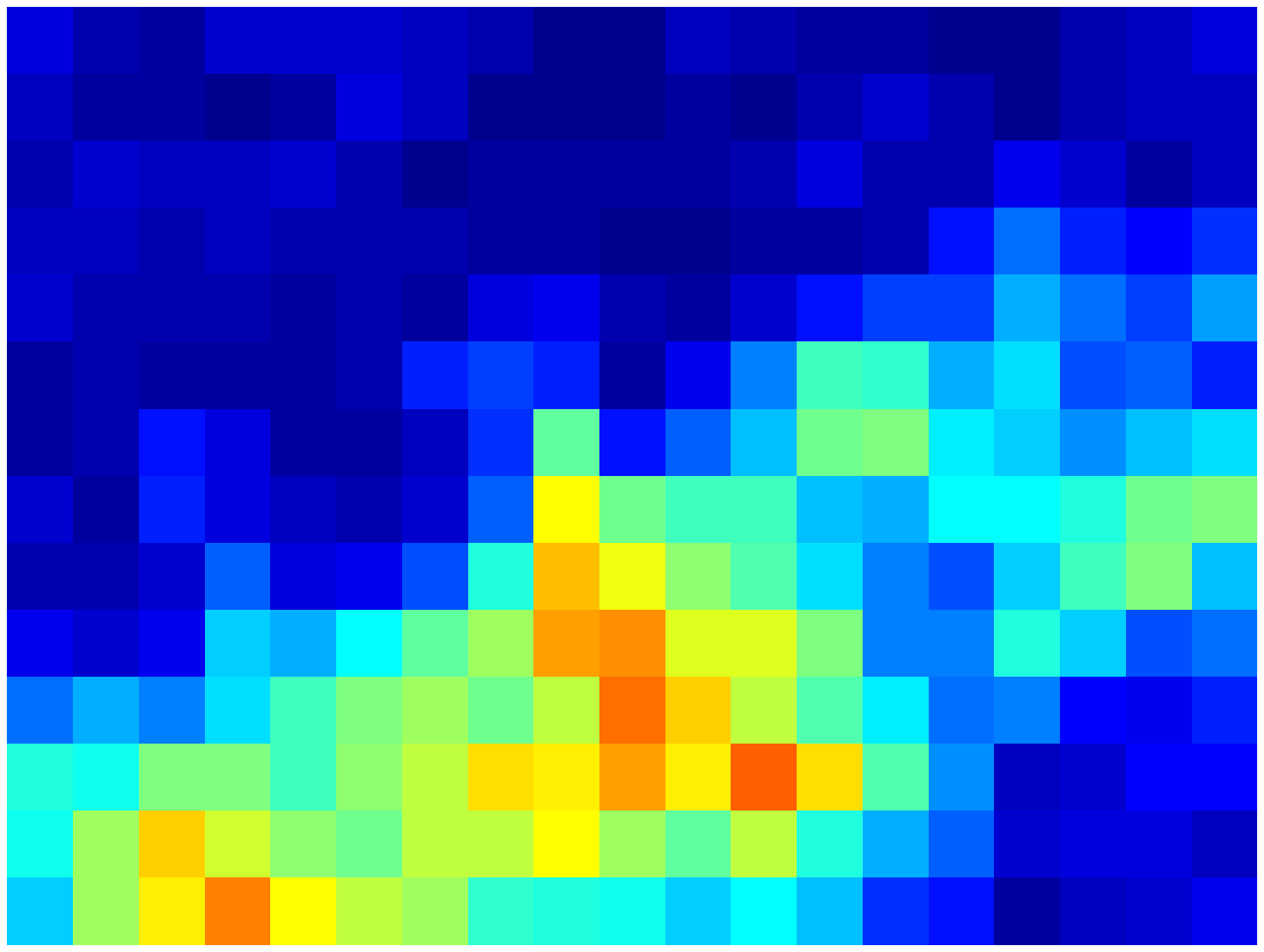} \\
            \includegraphics[width=3.5cm,height=2.8cm]{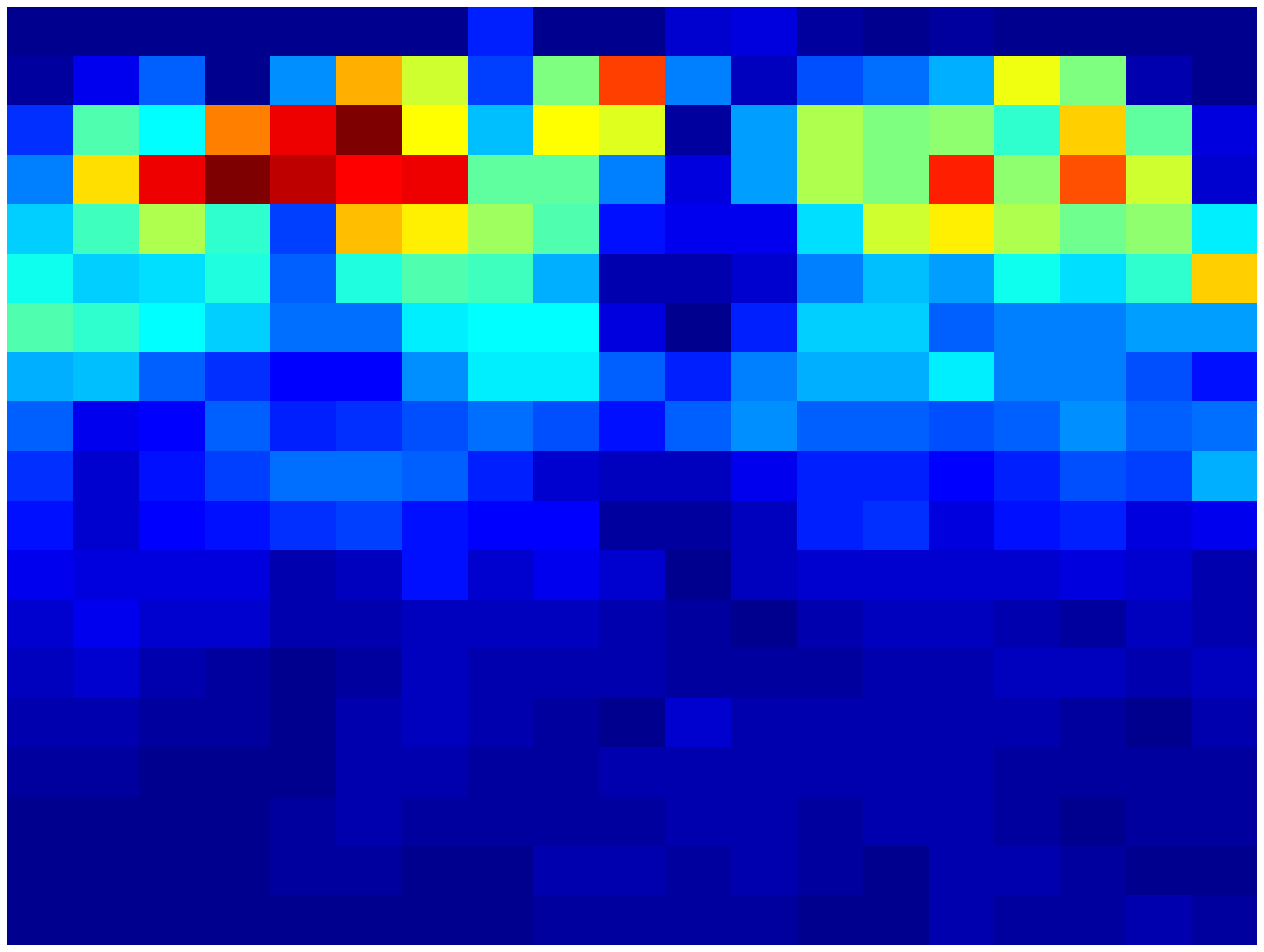}
        \end{tabular}
    }
    \subfloat[After MRF]{
        \begin{tabular}{@{}c@{}}
            \includegraphics[width=3.5cm,height=2.8cm]{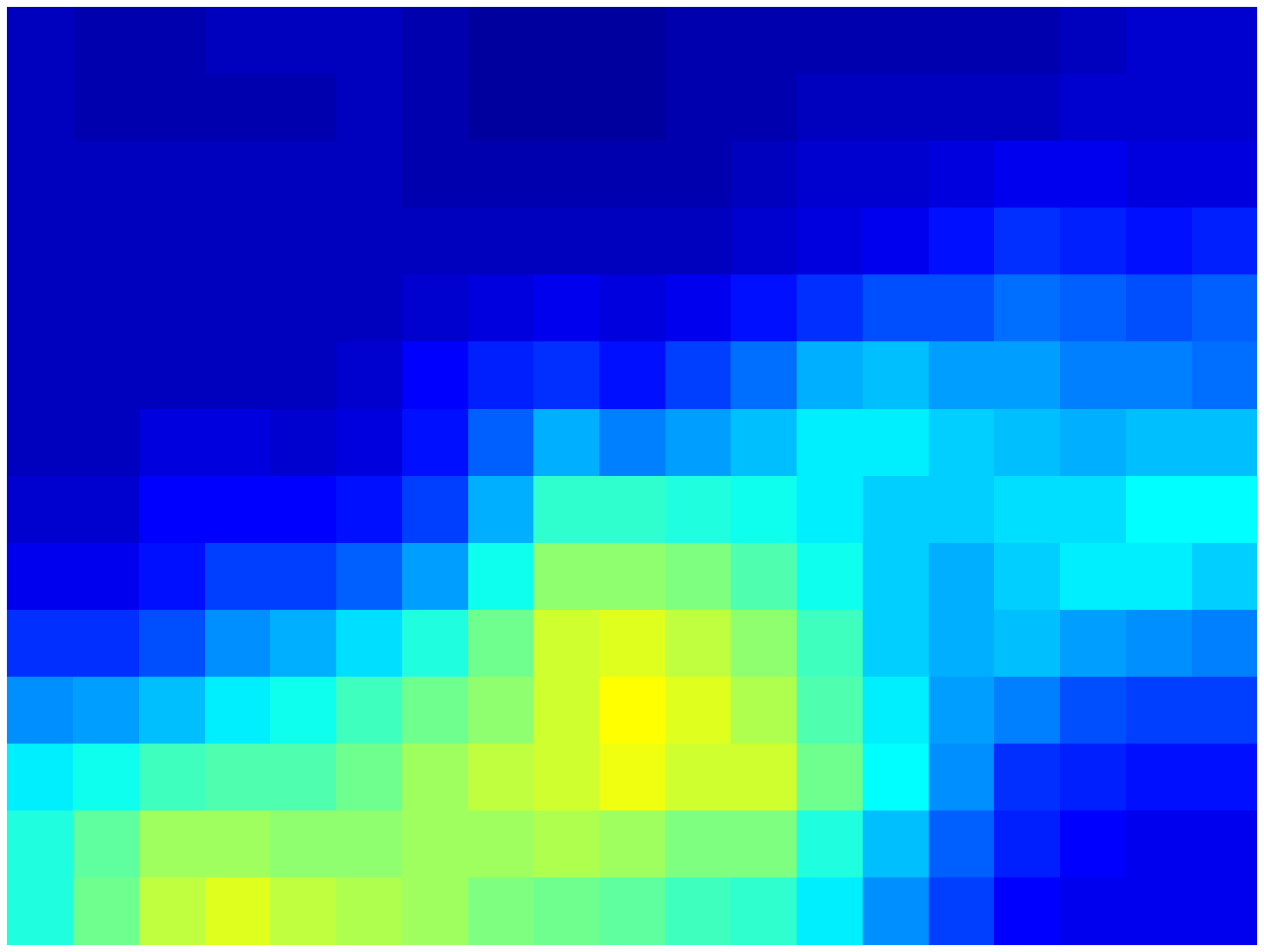} \\
            \includegraphics[width=3.5cm,height=2.8cm]{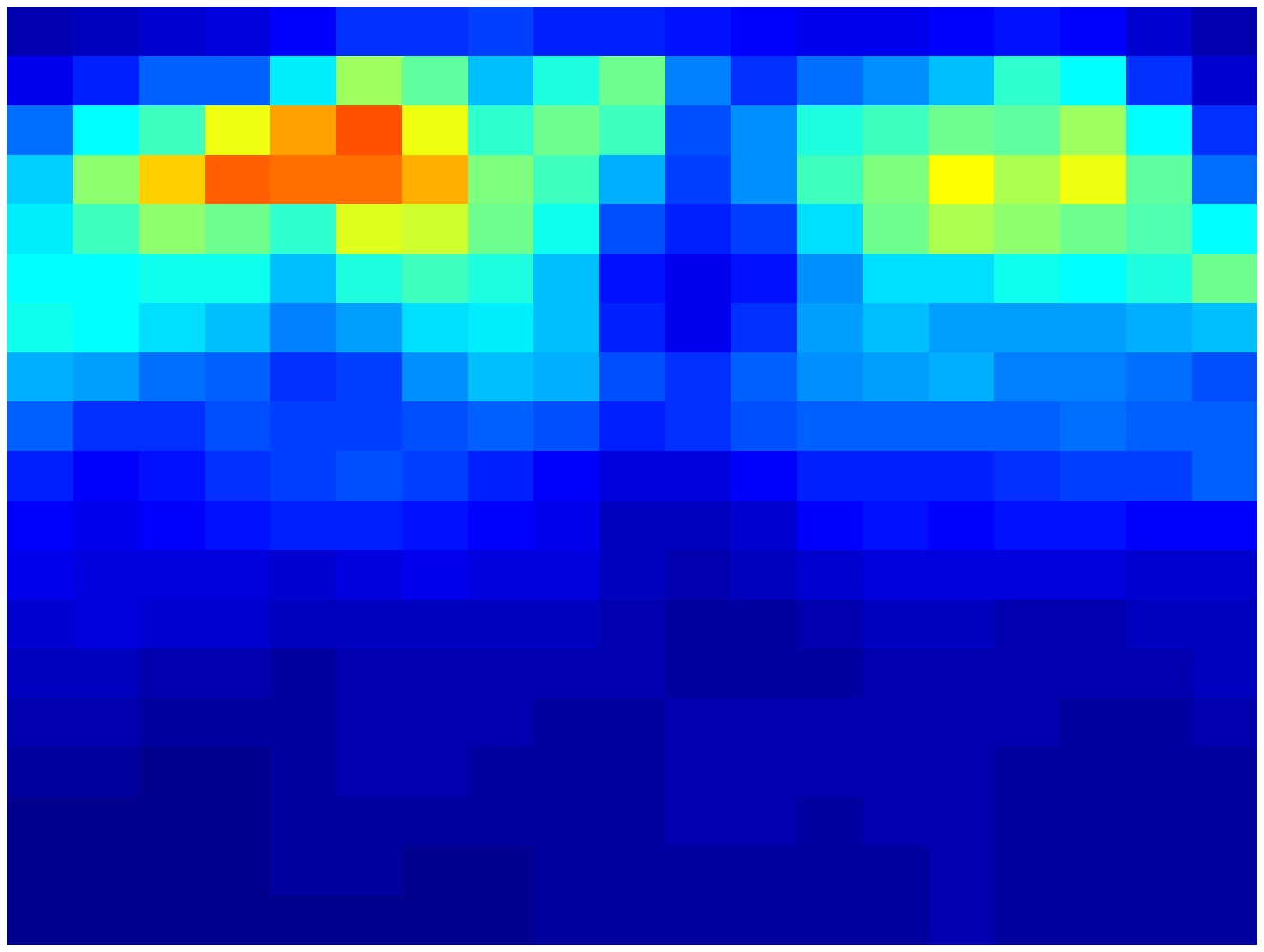}
        \end{tabular}
    }
    \caption{Smoothing of adjacent local counts by MRF. The first column are crowd images, the second, third, and final columns are the ground truth density map, the estimated density map, and the density map smoothed by the MRF, respectively.}
    \label{fig:MRF}
\end{figure*}

For the problem of smoothing the adjacent patches count, \(D_p(c_p)\) and \(V(c_p - c_q)\) can take the form of the following functions:
\begin{equation}
    D_p(c_p) = \lambda min((I(p) - c_p)^2, DATA\_K)
\end{equation}
\begin{equation}
    V(c_p - c_q) = min((c_p - c_q)^2, DISC\_K)
\end{equation}
where \(\lambda\) is a weight of the energy items, $I(p)$ is the ground truth count of the patch $p$, \(DATA\_K\) and \(DISC\_K\) are the truncating item of \(D_p(c_p)\) and \(V(c_p - c_q)\), respectively. The truncating item makes the cost function stop growing after the difference becomes large, which allows for large discontinuities. The above energy function minimization problem can be efficiently solved by belief propagation algorithm \cite{felzenszwalb2006efficient}. Fig. \ref{fig:MRF} shows the smoothing effect of the MRF on the local counting results. We can see that the density map is closer to the ground truth after smoothed by the MRF.

\section{Experiments}
We evaluate our method on UCF and Shanghaitech crowd counting datasets. Each image is divided with patch size 100 x 100 pixels and stride size 50 pixels. The final count of the whole image is obtained by calculating the sum of the count of all non-overlapping patches. If the image patch is on the edge and its previous image patch has been summed, then only half count of this image patch will be summed. The proposed method is implemented in Matlab, and we utilize MatConvNet \cite{vedaldi2015matconvnet}, a Matlab toolbox implementing CNN for computer vision applications, which provides many pre-trained CNN model for image classification, segmentation, face recognition and text detection. \par
\indent We utilize two evaluation criteria: the mean absolute error (MAE) and the mean squared error (MSE), which are defined as follows:
\begin{equation}
    MAE = \frac{1}{N}\sum_{i=1}^{N}|g_i - e_i|
\end{equation}
\begin{equation}
    MSE = \sqrt{\frac{1}{N}\sum_{i=1}^{N}(g_i - e_i)^2}
\end{equation}
where $N$ is the number of test images and \(g_i\) and \(e_i\) are the ground truth and the estimate count of the \(i\)-th image, respectively.

\subsection{UCF dataset}
The UCF dataset \cite{idrees2013multi} is a very challenging dataset because the scene of the each image are different and the crowd count of the image changes dramatically. More specifically, there are 50 images with counts ranging between 94 and 4543 with an average of 1280 individuals per image. The ground truth positions of individuals are marked by the authors and there is a total of 63705 annotations in the 50 images. 

\begin{table}[ht]
    \centering
    \caption{Comparing results of different methods on the UCF dataset.}
    \begin{tabular}{|m{3.5cm}<{\centering}|m{1.5cm}<{\centering}|m{1.5cm}<{\centering}|}
        \hline
        \textbf{Method} & \textbf{MAE} & \textbf{MSE} \\ 
        \hline
        Rodriguez et al. \cite{rodriguez2011density} & 655.7 & 697.8 \\ 
        \hline
        Lempitsky et al. \cite{lempitsky2010learning} & 493.4 & 487.1 \\
        \hline
        Idrees et al. \cite{idrees2013multi}  & 419.5 & 541.6\\
        \hline
        Zhang et al. \cite{zhang2015cross} & 467.0 & 498.5 \\
        \hline
        MCNN \cite{zhang2016single} & 295.1 & 490.2 \\
        \hline
        Shang et al. \cite{shang2016end} & 270.3 &  - \\
        \hline
        \textbf{Proposed} & \textbf{254.1} & \textbf{352.5} \\
        \hline
    \end{tabular}
    \label{tab:UCF}
\end{table}

\begin{figure}[H]
    \centering
    \includegraphics[width=0.60\linewidth]{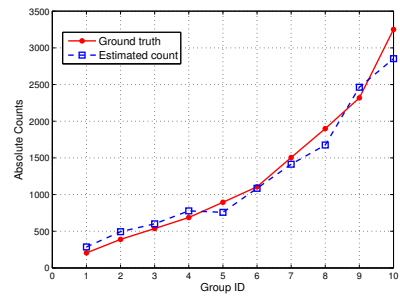}
    \caption{The comparison of the ground truth and the estimated count on UCF dataset. Absolute counts in the vertical axis is the average crowd number of images in each group.}
    \label{fig:UCF_group}
\end{figure}
Followed by Idrees et al. [6], we use 5-fold cross-validation to test the performance of our algorithm. Since the image of the UCF dataset is gray, we extend the image to three channels by copying the data. The counting result of our method and the comparison with other methods can be viewed in Table \ref{tab:UCF}. The experimental results of other methods come from their papers and the same for Shanghaitech\cite{zhang2016single} dataset. We can see that our proposed CNN-MRF outperforms the other methods, including the state of the art methods MCNN \cite{zhang2016single} and Shang et al. \cite{shang2016end}. In Fig \ref{fig:UCF_group}, we compare the estimated count with the ground truth in more details. The images are divided into 10 groups according to crowd counts in an increasing order. The divide-count-sum strategy narrows the range of the crowd count so that the proposed method can give an accurate estimate of the total counts of the images at all ranges.

\subsection{Shanghaitech dataset}
Shanghaitech dataset \cite{zhang2016single} is a large-scale crowd counting dataset which contains 330,165 people heads annotation in 1,198 images. The dataset consists of two parts, Part\textunderscore A is collected from the Internet and Part\textunderscore B is taken from the busy streets of metropolitan areas in Shanghai. The average crowd count of the Part\textunderscore A is 501.4, and the average number of the Part\textunderscore B dataset is 123.6. The crowd density of Part\textunderscore A is significantly larger than that in Part \textunderscore B. This dataset has been randomly divided into training and testing: 300 images of Part\textunderscore A are used for training and the remaining 182 images for testing, and 400 images of Part\textunderscore B are for training and 316 images for testing. Table \ref{tab:Shanghaitech} reports the results of different methods in the two parts. LBP+RR \cite{zhang2016single} is a regression based method which uses Local Binary Pattern (LBP) features extracted from the original image as input and uses ridge regression (RR) to predict the crowd number for each image. Our method significantly outperforms state-of-the-art methods.

\begin{table}[ht]
    \centering
    \caption{Comparing results of different methods on the Shanghaitech dataset.}
    \begin{tabular}{|c|c|c|c|c|}
        \hline
        \multirow{2}{*}{\textbf{Method}} & \multicolumn{2}{c|}{\textbf{Part\textunderscore A}} & \multicolumn{2}{c|}{\textbf{Part\textunderscore B}} \\
        \cline{2-5}
        & \textbf{MAE} & \textbf{MSE} & \textbf{MAE} & \textbf{MSE} \\
        \hline
        LBP+RR \cite{zhang2016single} & 303.2 & 371.0 & 59.1 & 81.7 \\
        \hline
        Zhang et al. \cite{zhang2015cross} & 181.8 & 277.7 & 32.0 & 49.8 \\
        \hline
        MCNN \cite{zhang2016single} & 110.2 & 173.2 & 26.4 & 41.3 \\
        \hline
        \textbf{Proposed} & \textbf{79.1} & \textbf{130.1} & \textbf{17.8} & \textbf{26.0}\\
        \hline
    \end{tabular}
    \label{tab:Shanghaitech}
\end{table}

\begin{figure}[ht]
    \centering
    \subfloat[Part\_A]{\includegraphics[width=0.45\linewidth]{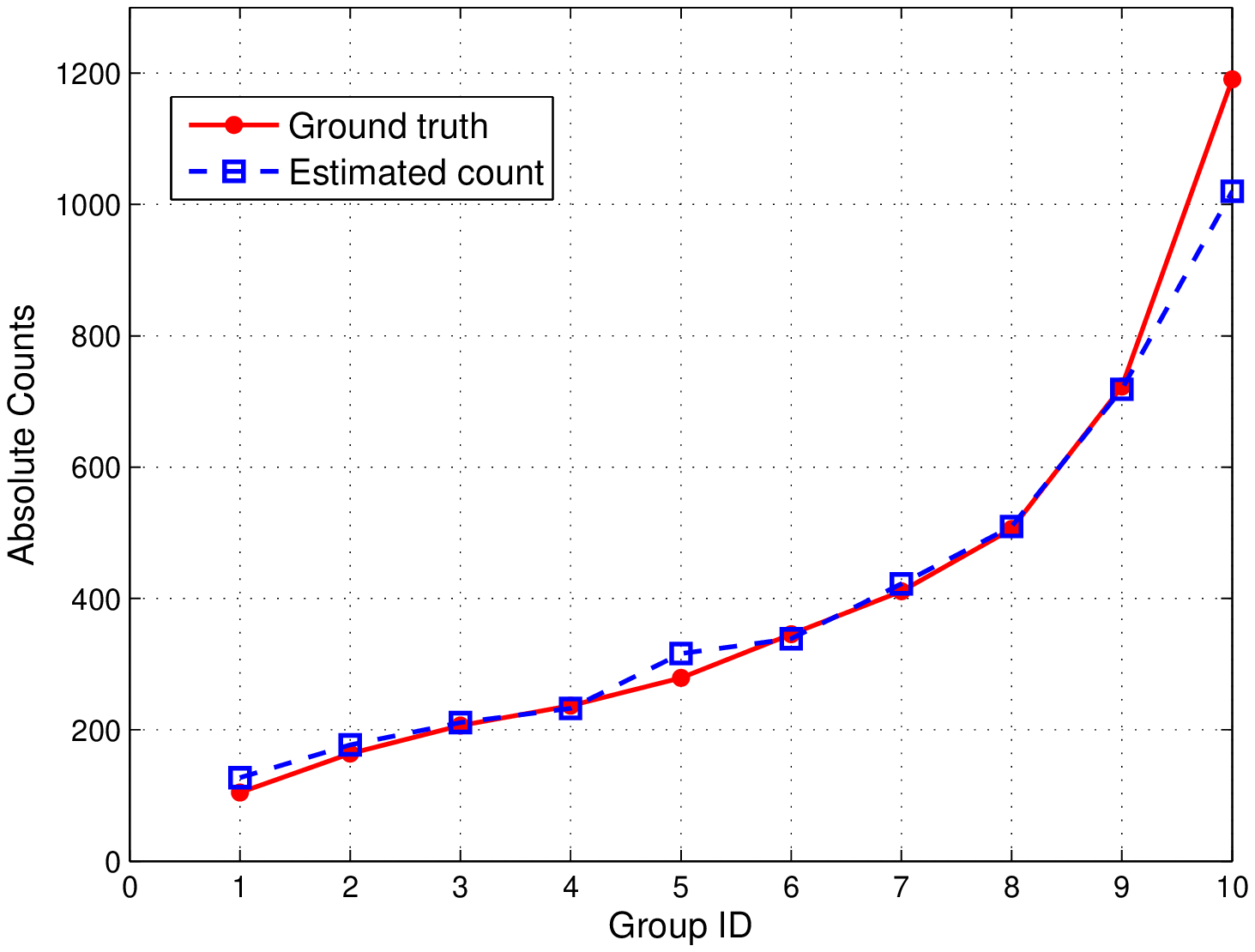}}
    \subfloat[Part\_B]{\includegraphics[width=0.45\linewidth]{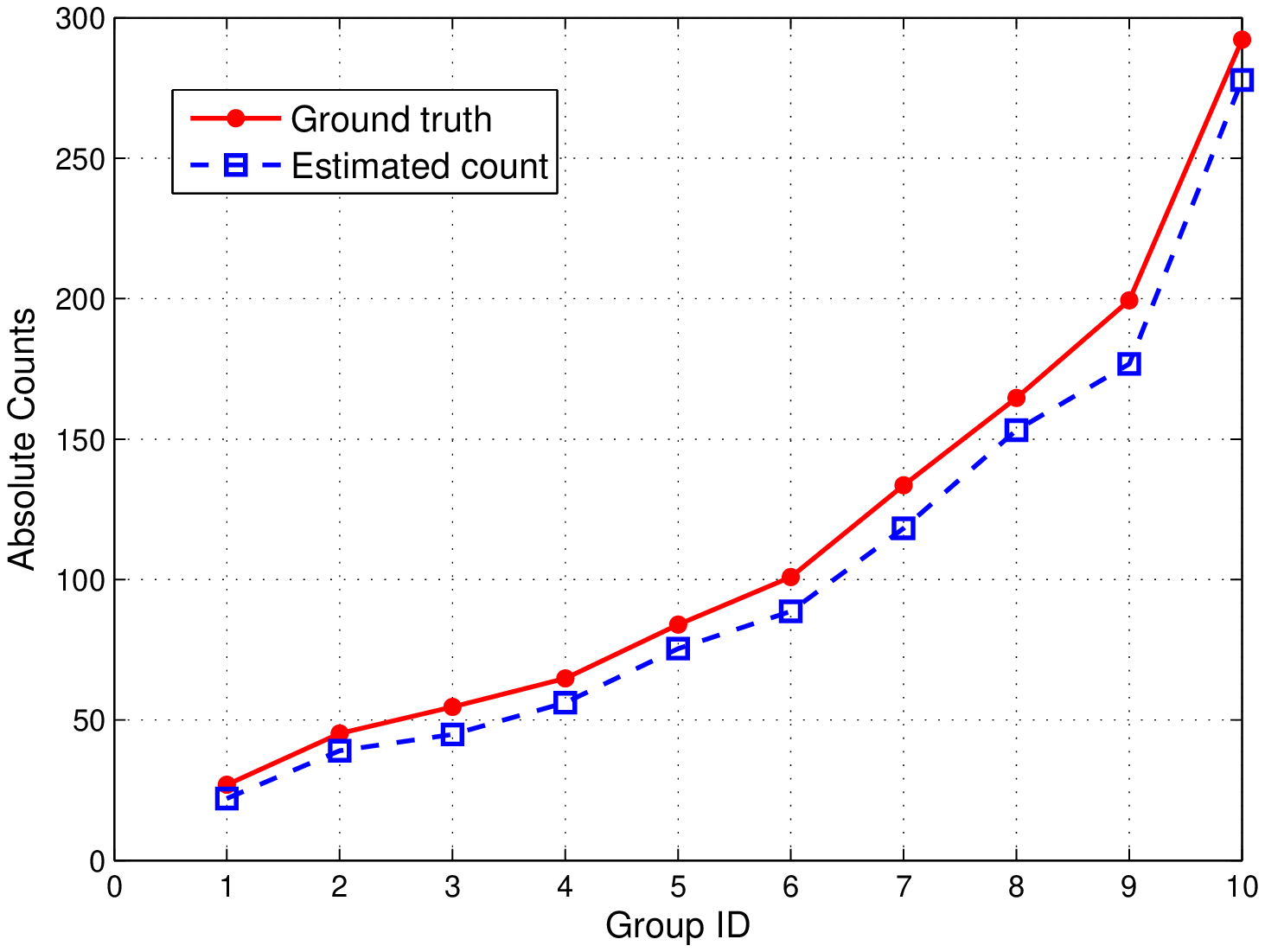}}
    \caption{The comparison of the ground truth and the estimated count on Shanghaitech dataset.}
    \label{fig:Shanghaitech_group}
\end{figure}

\begin{figure}[ht]
    \centering
    \subfloat[GT: 2391 C: 2362]{\includegraphics[width=3.5cm,height=2.5cm]{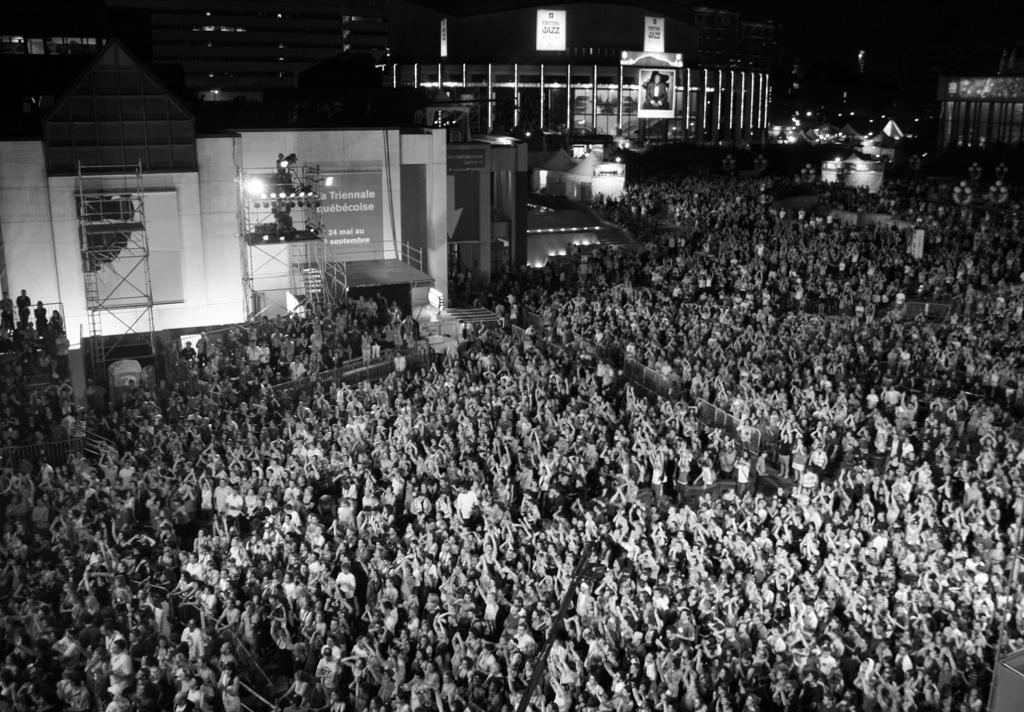}} \quad
    \subfloat[GT: 1601 C: 1632]{\includegraphics[width=3.5cm,height=2.5cm]{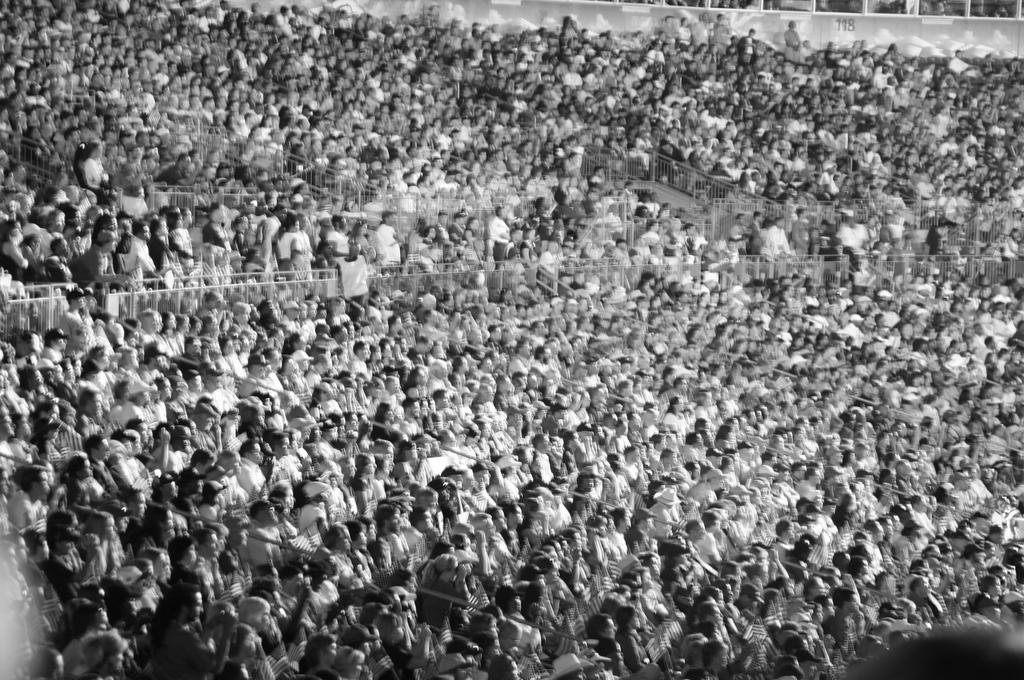}} \quad
    \subfloat[GT: 416  C: 414]{\includegraphics[width=3.5cm,height=2.5cm]{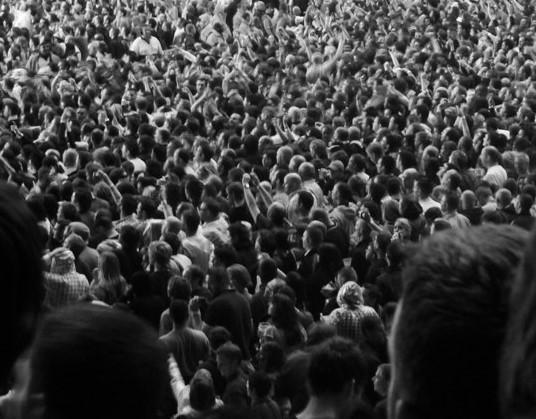}} \quad
    \subfloat[GT: 91   C: 91]{\includegraphics[width=3.5cm,height=2.5cm]{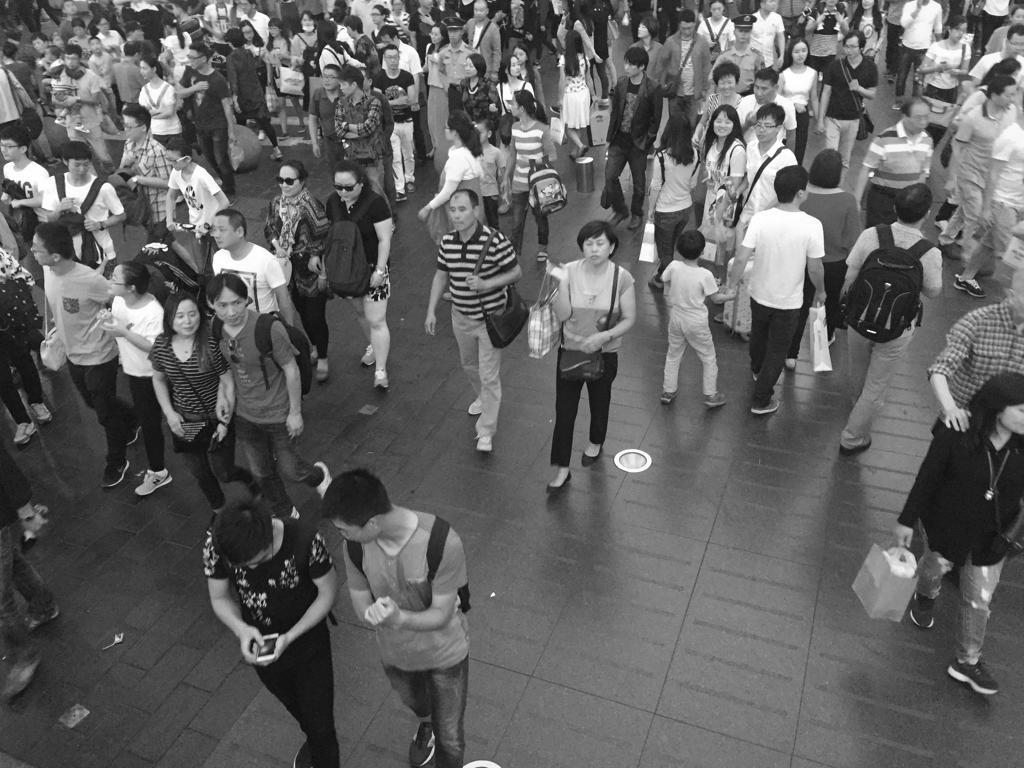}}
    \caption{Sample images with their respective ground truth and estimated count. GT is ground truth of the image, and C is the predicted count by the proposed method. Image (a) and (b) from the UCF dataset, (c) and (d) from the Shanghaitech dataset.}
    \label{fig:result}
\end{figure}

Similar to Fig \ref{fig:UCF_group}, the comparison of the ground truth and the estimated count on Shanghaitech dataset can be seen in Fig \ref{fig:Shanghaitech_group}. We can see that the proposed method can estimate the crowd count accurately and is robust to the large variation in crowd density. Some counting examples of the images with the associated ground truth counts can be seen in Fig \ref{fig:result}.

\section{Conclusion}
We present a CNN-MRF based approach to counting the crowd in a still image from different scenes. The features extracted from the CNN model trained for other computer vision tasks show a strong ability to represent crowd density. With the overlapping patches divided strategies, the adjacent local counts are highly correlated. This correlation can be used by the MRF to smooth the adjacent local counts to obtain a more accurate overall count. Experimental results demonstrate that the proposed approach achieve superior performance compared with several recent related methods.

\acknowledgements
This research was partially supported by the National Nature Science Foundation of China (No.61373084).

\bibliography{crowd}
\bibliographystyle{jaciiibibtex}

\end{document}